\documentclass{article}

\usepackage{natbib}
\usepackage{graphicx}
\usepackage[english]{babel}
    \usepackage[preprint]{sd_mad}

\usepackage{amsmath}
\usepackage[utf8]{inputenc} 
\usepackage[T1]{fontenc}    
\usepackage{hyperref}       

\usepackage{url}            
\usepackage{booktabs}       
\usepackage{amsfonts}       
\usepackage{nicefrac}       
\usepackage{microtype}      
\usepackage{xcolor}         
\usepackage{amssymb}
\usepackage{multirow}
\usepackage{wrapfig}
\usepackage{subcaption}
\usepackage{float}
\usepackage{graphicx}
\usepackage{enumitem}

\newtheorem{theorem}{Theorem}[section]

\newtheorem{definition}[theorem]{Definition}

\newtheorem{remark}[theorem]{Remark}

\title{SD-MAD: Sign-Driven Few-shot Multi-Anomaly Detection in Medical Images}

\author{%
  Kaiyu Guo\thanks{
  Equal contribution. This research was conducted during an internship at the Shanghai Academy of Artificial Intelligence for Science.
  }\\
Shanghai Academy of AI for Science \\
University of Queensland\\
  Brisbane, Australia \\
  \And
  Tan Pan\footnotemark[1]\\
    Fudan University\\
    Shanghai Academy of AI for Science \\
  Shanghai, China \\
  \AND
  Chen Jiang\thanks{Correspoding author} \\
  Shanghai Academy of AI for Science \\
  Shanghai, China \\
  \And
  Zijian Wang \\
  University of Queensland\\
  Brisbane, Australia \\
  \And
  Brian C. Lovell \\
  University of Queensland\\
  Brisbane, Australia \\
  \And
  Limei Han\\
  Fudan University\\
  Shanghai Academy of AI for Science \\
  Shanghai, China \\
  \And
  Yuan Cheng\footnotemark[2]\\
   Fudan University\\
   Shanghai Academy of AI for Science \\
  Shanghai, China \\
  \And
  Mahsa Baktashmotlagh\\
  University of Queensland\\
  Brisbane, Australia \\
}

\begin{document}

\maketitle

\begin{abstract}
Medical anomaly detection (AD) is crucial for early clinical intervention, yet it faces challenges due to limited access to high-quality medical imaging data, caused by privacy concerns and data silos. Few-shot learning has emerged as a promising approach to alleviate these limitations by leveraging the large-scale prior knowledge embedded in vision-language models (VLMs). Recent advancements in few-shot medical AD have treated normal and abnormal cases as a one-class classification problem, often overlooking the distinction among multiple anomaly categories. Thus, in this paper, we propose a framework tailored for few-shot medical anomaly detection in the scenario where the identification of multiple anomaly categories is required.  To capture the detailed radiological signs of medical anomaly categories, our framework incorporates diverse textual descriptions for each category generated by a Large-Language model, under the assumption that different anomalies in medical images may share common radiological signs in each category. Specifically, we introduce SD-MAD, a two-stage \textbf{S}ign-\textbf{D}riven few-shot \textbf{M}ulti-\textbf{A}nomaly \textbf{D}etection framework: (i) Radiological signs are aligned with anomaly categories by amplifying inter-anomaly discrepancy; (ii) Aligned signs are selected further to mitigate the effect of the under-fitting and uncertain-sample issue caused by limited medical data, employing an automatic sign selection strategy at inference.  Moreover, we propose three protocols to comprehensively quantify the performance of multi-anomaly detection. Extensive experiments illustrate the effectiveness of our method.  

\end{abstract}
\section{Introduction}
Medical anomaly detection (AD) has emerged as a critical area of research within the healthcare domain \cite{fernando2021deepmedicaladsurvey}. The detection of anomalies, such as tumors \cite{baid2021rsnabrainmri} and lesions \cite{ding2022unsupervisedbrainlesion}, is essential for prompt clinical intervention. However, access to  high-quality medical imaging data remains a significant challenge due to privacy concerns and institutional data silos, thereby highlighting the importance of few-shot learning approaches in medical anomaly detection.\par
\begin{figure}
    \centering
    \includegraphics[width=\linewidth]{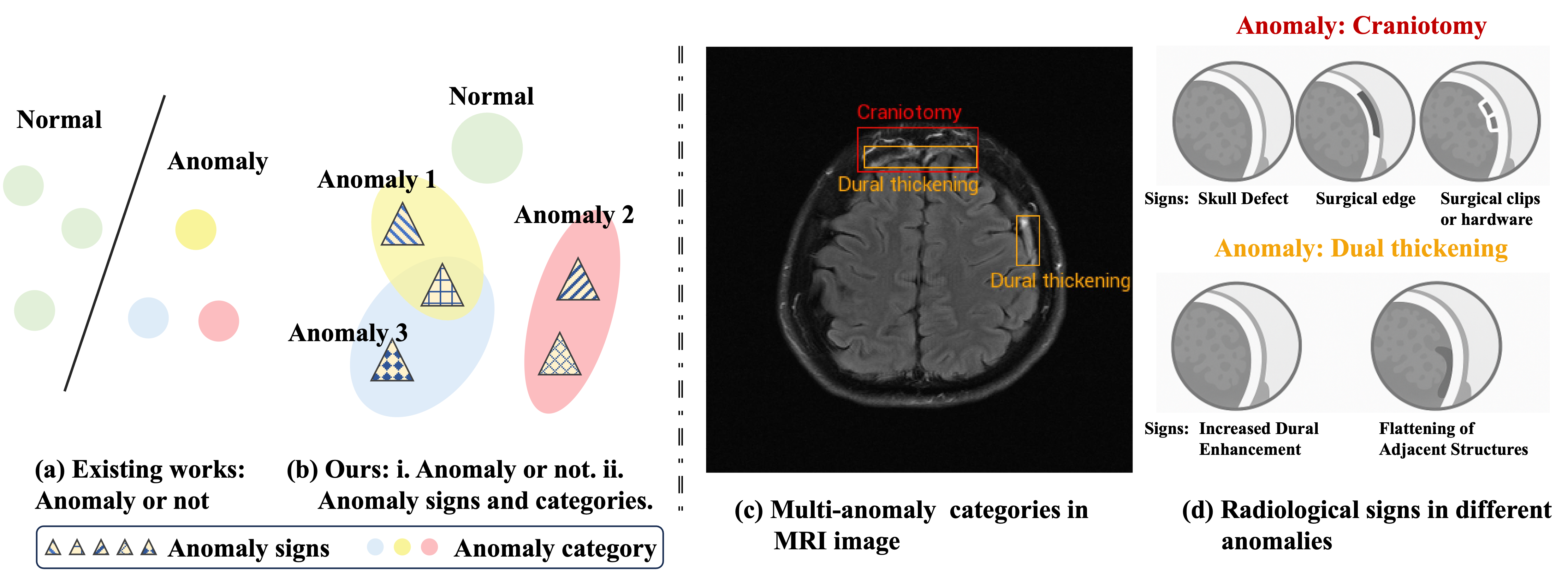}
    \caption{Figures (a) and (b) visualize the difference between our task and previous tasks. Figures (c) and (d) explain a multi-anomaly scenario, and radiological signs of different medical anomalies in the Brain MRI. }
    \label{fig:intro}
\end{figure}

Traditional few-shot anomaly detection \cite{sheynin2021hierarchicalfewshot,huang2022registrationfewshot} often struggles to generalize the model from the limited data to a universal situation because of the limited prior knowledge scale of the model. Recently, many works \cite{huang2024adaptingmvfa, AdaCLIP,gu2024univad} utilize the large-scale vision-language model (VLM), such as CLIP \cite{radford2021learningclip,wang2022medclip}, to help improve the generalization ability of the model in medical anomaly detection. Similar to traditional anomaly detection methods, these approaches identify anomalies by designing a score function that determines whether a given input is normal or abnormal (one-class classification). However, in real-world scenarios, especially in medical imaging, it is crucial to distinguish between different categories of anomalies, as they may correspond to varying pathological conditions and require distinct clinical responses. For example, distinguishing between a lung tumor and pneumonia in chest X-rays is crucial, as they require different treatment approaches: surgery or chemotherapy for cancer~\cite{montagne2021role}, and antibiotics for infection~\cite{bassetti2022new}. Thus, this paper aims to investigate scenarios involving the presence of diverse anomaly types by few-shot learning. The difference between the existing setting and our work is illustrated in Figure \ref{fig:intro}(a) and \ref{fig:intro}(b). \par

We hypothesize that different anomalies in medical images may share common radiological signs (\textit{e.g.}, ) in each category, such as abnormal density or shape, while also exhibiting unique signs that are specific to each anomaly category. These distinct features can provide valuable diagnostic information, enabling more accurate classification and treatment planning. By leveraging both shared and unique patterns, we aim to improve the detection and differentiation of various anomalies in medical imaging. Based on this hypothesis, firstly, we introduce a CLIP-based framework that explicitly \textbf{(i)} \textbf{links each anomaly class to a small set of textual “symptom” (signs) descriptions} and measures their similarity to image features. For each anomaly, we enumerate radiologic signs (e.g., “brain with craniotomy defect", “brain with unclear focal abnormality”) as prompts. As shown in Figure \ref{fig:intro}(d), aligning visual embeddings with these sign prompts allows the model to learn fine-grained inter-anomaly distinctions. However, recent work \cite{xia2024cares,wang2022medclip} reveals that prompt-based alignment in medical vision–language models can be uncertain: not all signs contribute equally, and some may even introduce noise in intra-class matching. To address this, at inference time, we \textbf{(ii) automatically select the most informative prompts for each few-shot example}~\cite{shum-etal-2023-automaticpromptselection}, thereby mitigating misleading matches within the same anomaly class. 
By addressing both inter-anomaly and intra-anomaly challenges, our approach delivers more accurate and reliable multi-anomaly detection under few-shot conditions. 

We structure the evaluation protocol for the multi-category medical AD task around three layers to capture the full spectrum of multi-anomaly detection performance: (1) assessing the model's ability to distinguish between normal and abnormal instances; (2) evaluating the model’s ability to perform multi-label prediction across distinct anomaly types; and (3) assessing the model's ability to correctly identify the specific types of anomalies. Existing methods face challenges in adapting to the last two protocols, primarily because their scoring functions are not designed to generalize to these task settings. \par

\vspace{-4pt}

As summarized below, our contributions are threefold:
\begin{enumerate}
    \item \textbf{Framework for few-shot multi-anomaly detection.} We introduce a few-shot anomaly detector that natively handles multiple anomaly classes within a single model, based on learning the alignment of radiological signs and anomaly categories.   
    \item \textbf{Inter- and intra-anomaly alignment.} We align image embeddings with sets of anomaly-specific prompts during training and, at inference, automatically select the most informative prompts to mitigate the uncertain-sample issue in the vision–language alignment.  
    \item \textbf{Rigorous evaluation protocol.} We assess our approach on seven medical imaging datasets across three evaluation settings, covering both single-class and multi-anomaly scenarios, and demonstrate consistent improvements over state-of-the-art baselines.
\end{enumerate}
\section{Related Work}
\textbf{Medical Anomaly detection.} Traditional medical anomaly detection methods rely on well-curated anomaly datasets, training on normal images and evaluating on abnormal ones~\cite{bao2024bmad,cai2023dual,zhang2020viral,zhou2021proxy,xiang2023squid,hassanaly2024evaluation,linmans2024diffusion,graham2023latent}.  These approaches model the normal data distribution and identify anomalies as deviations from this distribution, achieving impressive performance. Many of these methods are designed for specific anatomical regions~\cite{ding2022unsupervised,xu2024afsc} and treat anomaly detection (AD) as a one-class classification problem~\cite{bao2024bmad,cai2023dual,jiang2023multi}. However, in real-world scenarios, the same individual may experience multiple diseases affecting the same organ. Recently, the open-set AD method~\cite{zhu2024anomalyopenset} has shifted focus to detecting multiple anomalies instead of relying on one-class classification. These methods require enough training data to formulate the expected distributions, which can be hard to adapt to few-shot setting. To address the challenge of limited large-scale labeled datasets, some approaches have explored few-shot anomaly detection techniques as follows.\par
\textbf{Few-shot Anomaly detection.} Few-shot anomaly detection has gained significant attention in recent years due to its ability to identify rare or unseen anomalies with limited labeled data. Previous models utilized disentangled representations of anomalies~\cite{ding2022catching} or contrastive learning mechanisms~\cite{yao2023explicit} to alleviate the bias, accounting for unseen anomalies.  MVFA~\cite{huang2024adapting} utilized multi-level adaptation and a contrastive framework to improve generalization across various medical datasets. UniVAD~\cite{gu2024univad} proposed a general framework to detect anomalies across different domains with a training-free unified model. AA-CLIP~\cite{ma2025aa} advanced CLIP model in a two-stage approach to enhance CLIP's anomaly discrimination ability. Although those methods perform well in various datasets, there is still a lack of few-shot multi-anomaly detection for medical data.  \par
\textbf{Vision-language model.} Vision-language models have demonstrated significant potential across a range of tasks. CLIP~\cite{radford2021learningclip} excels in image-text alignment and has been successfully applied to various applications, such as classification and text-image retrieval. To expand CLIP's capabilities to medical data, MedCLIP~\cite{wang2022medclip} was introduced as a foundation for medical image-text alignment. Based on those pre-trained foundation models, recent studies~\cite{hua2025hieclip,jin2025logicad,cao2025personalizing} in anomaly detection have leveraged pre-trained CLIP models for language-guided anomaly detection and segmentation, achieving impressive results and highlighting the promising potential of these models in this domain.
\section{Methodology}
In this section, we first formulate the problem of few-shot anomaly detection and few-shot multi-anomaly detection in medical images. Then we propose our methods within two parts: In section \ref{section3.2}, we propose a training method with a tailored adapter for vision-language models and an inter-anomaly representation learning loss function; In Section \ref{section3.3}, we propose an inference strategy to filter the outlier prompts, which aims to handle the intra-anomaly uncertain samples. Figure \ref{fig:pipeline} shows the overall pipeline of our model.

\subsection{Problem Formulation}
\label{sec:problemfrom}
\textbf{Few-shot medical anomaly detection}: Following the setting of previous work \cite{huang2024adaptingmvfa} on few-shot medical anomaly detection, the few-shot training samples can be presented as $\mathcal{D}_{few} = \{(x_i, c_i, s_i)\}^K_i$,

where $K$ is the number of samples, $x_i$ is the $i$-th image, the corresponding image-level label $c_i \in \{0,1\}$, and the pixel-level label $s_i\in \{0,1\}^{h\times w}$ is a binary mask with the same size $h\times w$ as the image $x_i$.
For a given test image $x_{test}$, image-level and pixel-level medical anomaly detection are evaluated with the corresponding image labels $c_{test}$ and pixel labels $s_{test}$.\par
\textbf{Few-shot medical anomaly detection with multiple anomaly categories}: Similar to the setting of few-shot medical anomaly detection,  few-shot training samples can be presented as $\mathcal{\hat{D}}_{few} = \{(x_i, \mathbf{c}_i)\}^K_i$, where $\mathbf{c}_i\in \{0,1\}^{d}$ is a $ d$-dimensional label. Since it is hard to access the pixel-level labels for the multi-anomaly medical datasets, we do not consider the pixel-level label in this setting. Thus, given a test image $x_{test}$, only image-level medical anomaly detection is evaluated with the corresponding image labels $\mathbf{c}_{test}$ in the scenarios where multiple anomaly categories exist.
\subsection{Training: Amplify Inter-anomaly Discrepancy} 
\label{section3.2}
\noindent\textbf{Shift Adapter.}
\begin{figure}
    \centering
\includegraphics[width=0.9\linewidth]{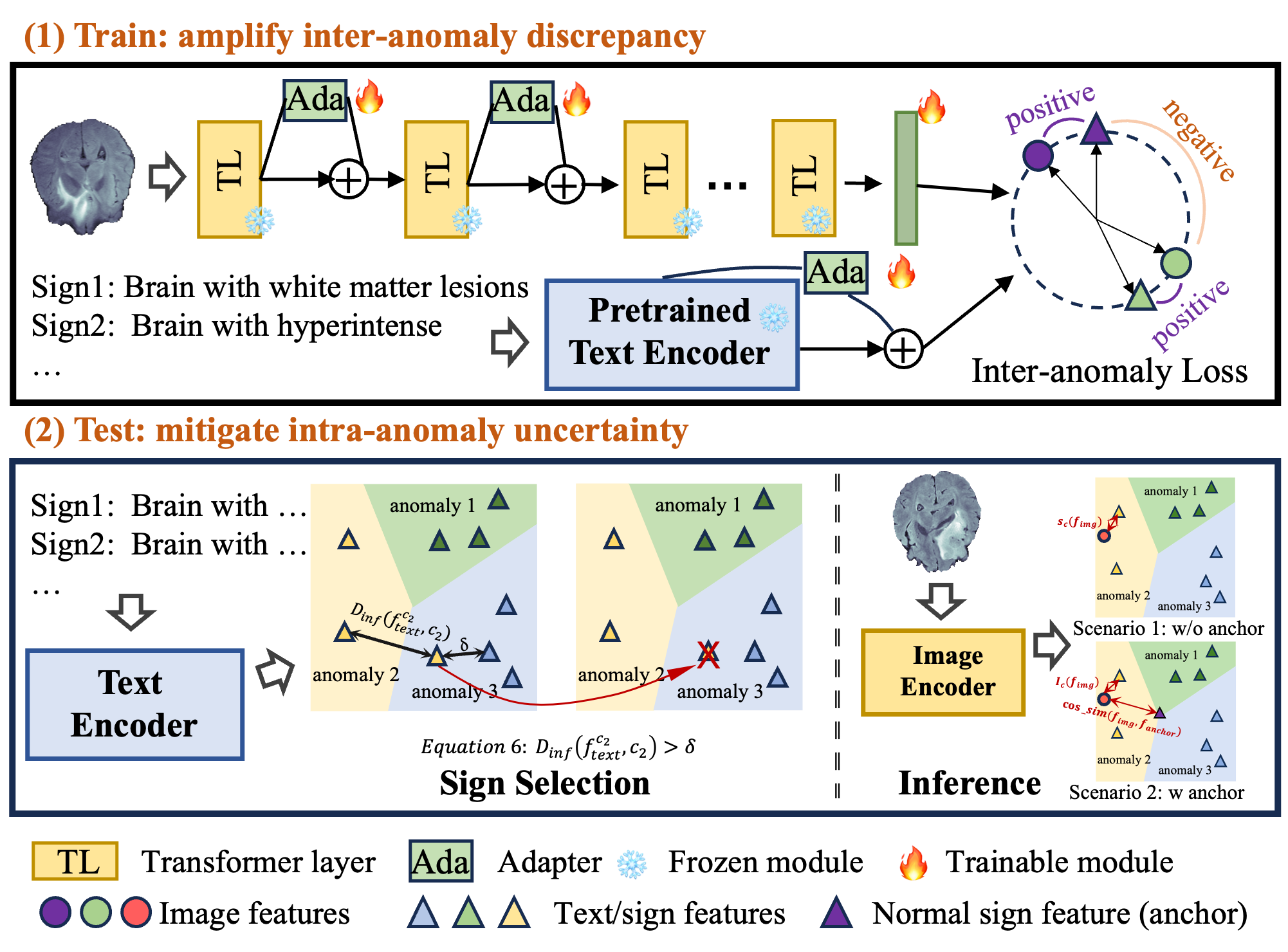}
    \caption{The pipeline of SD-MAD. In the framework, the training phase is designed to amplify inter-anomaly discrepancies, and the inference stage aims to handle the uncertain-sample problem in each anomaly category. }
    \label{fig:pipeline}
\end{figure}
To preserve the large-scale prior knowledge encoded in CLIP, we propose a shift adapter designed to effectively aggregate learning signals from few-shot samples while retaining the original prior information. The shift adapter is used for both image and text encoders, which is shown as our pipeline in Figure \ref{fig:pipeline}.\par
Considering the feature $\hat{f}_i^{in}$ is input of the adapter, which is also the output of the $i$-th transformer layer, the output of the adapter at the $i$-th transformer layer is 
\begin{equation}
    \hat{f}_i^{ada} = \alpha(W_i^2\alpha(W_i^1\hat{f}_i^{in})),
\end{equation}
where $W_i^1$ and $W_i^2$ are trainable linear weights of the adapter at the $i$-th transformer layer, $\alpha$ is the activation function. \par
Inspired from residual learning methods~\cite{he2016deepresnet}, we integrate the output of the original transformer layer $\hat{f}_i^{out}$ with $\hat{f}_i^{ada}$ by inner interpolation as follows:
\begin{equation}
    f_i^{out} = \lambda \hat{f}_i^{out} + (1-\lambda) \hat{f}_i^{ada},
\end{equation}
where $\lambda$ is the hyperparameter to control the interpolation ratio. To avoid the overfitting caused by the limited number of few-shot samples, we restrict the application of the adapter to four layers in the image encoder and one layer in the text encoder.\par
\noindent\textbf{Inter-anomaly Loss.} The text-vision alignment in CLIP depends on this contrastive learning insight with the cosine similarity. From the view of contrastive learning~\cite{oord2018representationinfonce, schroff2015facenet}, the distance of the positive image-text pairs should be smaller than the distance of negative text-image pairs. Towards this end, existing work directly minimize the cosine distance between the positive image-text pairs to align the text and image features as follows: 

\begin{equation}
    L_{img-text} = \min_{\theta}\sum_{i\in [1,N_c]} d(f_{img}^c, f_{text, i}^c).
    \label{loss:img-text}
\end{equation}

Here, $d(\cdot,\cdot)$ denotes the cosine distance between two input vectors, $\theta$ is the trainable parameters, image feature $f_{img}^c$ and detailed-description text features $f_{text, i}^c$ belong to anomaly category $c\in\mathcal{C}$ of the given image, $N_c$ is the number of text prompts corresponding to category $c$.

%

It is important to note that Equation \ref{loss:img-text} does not account for the distances of negative pairs. This is because simply increasing the distance between negative pairs provides limited utility in enabling the model to accurately identify the anomaly categories. For instance, given an abnormal image exhibiting only the anomaly of a lesion, the prediction may still fail despite strong alignment of positive pairs, as the model may erroneously assign high similarity scores to irrelevant categories, resulting in false positives. To handle this issue, we introduce an anchor feature $f_{anchor}$ that serves to define the boundary between normal and abnormal images. Thus, the following relationship should be satisfied. 
\begin{remark}
   Given an image feature belonging to category $c$, we have \[
   \sup_{i\in[1,N_c]} d(f_{img}^c, f_{text, i}^c) \leq d(f_{img}^c, f_{anchor}) \leq \inf_{k\neq c, j\in[1,N_k]} d(f_{img}^c, f_{text, j}^k)
   \]
\label{remark1}
\end{remark}
Given the image feature $f_{img}^c$ and category $c$, Remark \ref{remark1} indicates that $f_{anchor}$ serves as the hyperplane to separate the subspace of category $c$ and other categories. To distinguish the difference between the normal category and other anomalies simultaneously, we set the $f_{anchor}$ as the feature of the text prompt corresponding to normal images. 
According to Remark \ref{remark1}, we propose the following loss:
\begin{align}
    &\hat{d}_{positive,i}^c = \max(0,d(f_{img}^c, f_{text,i}^c)-d(f_{img}^c, f_{anchor})) \notag \\
    &\hat{d}_{negative,j}^{c,k} = \max(0,d(f_{img}^c, f_{anchor}^c)-d(f_{img}^c, f_{text,j}^k))\notag \\
    &L_{anchor} = \min_\theta \sum_{i\in[1,N_c]}\hat{d}_{positive,i}^c 
     + \sum_{k\neq c, j\in [1,N_k]}\hat{d}_{negative, j}^{c,k}
\end{align}
As discussed above, the overall loss for amplifying the inter-anomaly discrepancy is 
\begin{equation}
    L = L_{img-text} + L_{anchor}
\end{equation}
\subsection{Inference: Mitigate Intra-anomaly Uncertain-sample Issue}\label{section3.3}
During the inference stage, image features from the test set are evaluated against the text prompt features corresponding to each anomaly category. However, the limited number of few-shot training samples, combined with uncertainty in medical vision-language~\cite{xia2024cares}, may cause under-fitted features that fail to capture anomaly characteristic-specific information. Thus, to address this issue, we divided our inference stage into two parts as follows.\par 
\noindent\textbf{Sign Selection.} As we discussed above, each anomaly category contains several prompt features corresponding to the anomaly signs. Thus, there should be a labeling function $h_{text}(\cdot)$ satisfied $h_{text}(f_{text}^c) = c$. Therefore, we have the definition of the distance between given text feature $f_{text}$ and category $c$ in the following.
\begin{definition}
    Given a text feature $f_{text}$,
    the distance between the prompt feature $f_{text}$ and the decision region of the anomaly category $c$ is \[
    D_{inf}
    (f_{text},c) = \inf_{\{f'_{text}|h(f'_{text})=c, f'_{text}\neq f_{text}\}} d(f_{text}', f_{text})
    \] 
\label{def1}
\end{definition}
Definition \ref{def1} provides a definition of distance between the prompt feature $f_{text}$ and the decision region $\{f'_{text}|h(f_{text}') = c)\}$. For the ideal situation, we have the following relation. 
\begin{remark}
    Given a text feature $f_{text}^c$ belonging to category $c$, we have 
    \[
    D_{inf}(f_{text}^c,c) < \delta 
    \]
    Where $\delta \triangleq \inf_{k\neq c, k\in\mathcal{C}} D_{inf}(f_{text}^c,k)$
    \label{remark2}
\end{remark}
As shown in Figure \ref{fig:pipeline}, the outlier text features in each category may break the relation in Remark \ref{remark2}. However, in the inference time, the text features are fixed. Thus, we propose to modify the labeling function $h(\cdot)$ to mitigate this problem. \par
Given a text feature $f_{text}^c$ which satisfies $h(f_{text}^c) =c $, the new labeling function is defined as 
\begin{equation}
h_{new}(f_{text}^c) = \left\{
\begin{array}{cc}
     c& \textrm{if $D_{inf}(f_{text}^c, c)<\delta$} \\
     -1& \textrm{else}
\end{array}\right.
\label{eq: newlabel}
\end{equation}
 The Equation \ref{eq: newlabel} indicates that the new labeling function $h_{new}(\cdot)$ discards the distorted features that break the relation in Remark \ref{remark2} for each anomaly category. The sign selection process can be viewed in Fig. ~\ref{fig:pipeline} (2). This labeling function is used for the score function design, which we will discuss in the following.

\noindent\textbf{Inference.} Unlike previous methods~\cite{huang2024adaptingmvfa, jeong2023winclip}, which focus solely on evaluating the Area Under the Receiver Operating Characteristic curve (AUROC) using a continuous scoring function, we additionally consider scenarios that require binary predictions. For the binary prediction, the anchor feature is required for the evaluation. \par
\textbf{Scenario 1: Continuous scoring function without anchor feature.} Without anchor feature, for the given category $c$ and the image feature $f_{img}$, the score function corresponding to $c$ is 
\begin{equation}
    s_c(f_{img}) = \sup_{h_{new}(f_{text})=c} cosine\_similarity(f_{img}, f_{text}).
\end{equation}
As we discussed in Section \ref{sec:problemfrom}, the label of image $x$ is a vector $\mathbf{c}$. Thus, the score vector corresponding to $\mathbf{c}$ is $\mathbf{s_c} = \{s_{c_i}(f_{img})\}_{i=1}^K $, where $K$ is the number of anomaly categories.  \par
\textbf{Scenario 2: Binary prediction with anchor feature.} With the anchor feature, we can achieve the binary prediction for each anomaly category. The prediction $p_c$ for category $c$ with a give image feature $f_{img}$ is 
\begin{equation}
    p_c(f_{img}) = \left\{
\begin{array}{cc}
     1& \textrm{if $I_c(f_{img})>cosine\_similarity(f_{img},f_{anchor}) $} \\
     0& \textrm{else}
\end{array}\right.
\end{equation},
where $I_c(f_{img}) \triangleq\inf_{h_{new}(f_{text})=c} cosine\_sim(f_{img}, f_{text})$. The precision vector corresponding to $\mathbf{c}$ is  $\mathbf{p_c} = \{p_{c_i}(f_{img})\}_{i=1}^K $. This prediction can be used for the evaluation with the Hamming score and the subset accuracy score. 
\section{Experiments}
\subsection{Experimental Setup}
\textbf{Evaluation protocols} We introduce three evaluation protocols: 1) for general anomaly detection, following previous works~\cite{huang2024adaptingmvfa,jeong2023winclip}, we quantify the performance with area under the receiver operating curve (AUROC) metric on image- and pixel-level; 2)We introduce Hamming score and subset accuracy to evaluate the performance on multi-label prediction on the task of multi-anomaly detection; 3) We exploit the AUROC metric for each class to evaluate the performance on the specific types of anomalies. Specifically, given the anomaly type $c$, the binary label is set as 1 for the images belonging to type $c$, and 0 for the others. 

\textbf{Dataset} We evaluate the methods with 7 datasets. For general medical anomaly detection, we follow the BMAD benchmark \cite{bao2024bmad}, which includes 6 datasets: Brain MRI~\cite{baid2021rsnabrainmri,bakas2017advancingbrainmri, menze2014multimodalbrats}, Liver CT~\cite{bilic2023liverctlist,landman2015miccailiverct}, retinal OCT~\cite{kermany2018identifyingoct17, hu2019automatedresc}, Chest X-ray~\cite{wang2017hospitalchestxray}, and Digital Histopathology~\cite{bejnordi2017diagnostichis}. Among these datasets, both image- and pixel-level metrics are evaluated for BrainMRI~\cite{baid2021rsnabrainmri,bakas2017advancingbrainmri,menze2014multimodalbrats}, LiverCT~\cite{bilic2023liverctlist,landman2015miccailiverct}, and RESC~\cite{hu2019automatedresc}. For the other datasets, namely OCT17~\cite{kermany2018identifyingoct17}, ChestXray~\cite{wang2017hospitalchestxray} and HIS~\cite{bejnordi2017diagnostichis}, only image-level scores are evaluated. \par
The experiments for multi-anomaly detection are built from the brain MRI dataset in fastMRI+~\cite{zhao2022fastmri+,zbontar2018fastMRI}. We select 6 anomaly categories and the same slice-level images, namely slice 0, 5 and 10, for the multi-anomaly detection tasks. More details can be viewed in the Appendix. \par
\textbf{Training details} We select CLIP with ViT-L/14~\cite{dosovitskiy2020imagevit} as the backbone model with the size of input as 240$\times$240. We employ our shift adapter to the 6-,8-,18- and 24-th layers in the transformer of the CLIP image encoder and to the last layer to the transformer of the CLIP text encoder. Every training process is conducted in 50 epochs. The training process requires 4000 Mib GPU memory for the model. The experiments are conducted on an A100 GPU.
\subsection{General Few-shot Medical Anomaly Detection}
We first evaluate our method under the setting of general few-shot anomaly detection. We conduct the 4-shot experiments with state-of-the-art few-shot medical anomaly detection methods, MVFA~\cite{huang2024adaptingmvfa}, and other few-shot anomaly detection methods, namely  BRA~\cite{ding2022catchingbra} and BGAD~\cite{yao2023explicitbgad}. To adapt our method to pixel-level score, we combine our method with MVFA. Specifically, we aggregate our inter-anomaly loss with the losses of MVFA.

As Table \ref{tab:general} shows, even though our methods are not designed for the general few-shot anomaly detection, we still average outperform other methods. In addition, for LiverCT, our method can significantly improve the anomaly detection performance. 
\begin{table}[t]
\centering
\caption{Comparison on general anomaly detection. "Avg." is short for "average".}
\vspace{1ex}
\begin{tabular}{llcccc}
\toprule
 & \textbf{Dataset} & \textbf{DRA} & \textbf{BGAD} & \textbf{MVFA} & \textbf{Ours} \\
\midrule
\multirow{6}{*}{\begin{tabular}[c]{@{}l@{}}Img-level\\ (AUROC(\%))\end{tabular}} 
& BrainMRI   & 80.6 & 83.6 & 92.4 & 91.4 \\
& LiverCT    & 59.6 & 72.5 & 81.2 & 86.9 \\
& RESC       & 90.9 & 86.2 & 96.2 & 95.2 \\
& HIS        & 68.7 & -    & 82.7 & 81.6 \\
& ChestXRay  & 75.8 & -    & 82.0 & 82.7 \\
& OCT        & 99.0 & -    & 99.4 & {99.8} \\
\midrule
\multirow{3}{*}{\begin{tabular}[c]{@{}l@{}}Pixel-level\\ (AUROC(\%))\end{tabular}} 
& BrainMRI   & 74.8 & 92.7 & 97.3 & {96.5} \\
& LiverCT    & 71.8 & 98.9 & {99.7} & 99.5 \\
& RESC       & 77.3 & 93.8 & 99.0 & {99.0} \\\hline
& Avg. & 77.6 & 88.0 & 92.2& \textbf{92.5}\\
\bottomrule
\end{tabular}
\vspace{-2ex}
\label{tab:general}
\end{table}

\subsection{Multi-category Few-shot Medical Anomaly detection}
As discussed above, we introduce two evaluation protocols for multi-category few-shot medical anomaly detection: evaluating the model’s ability to perform multi-label prediction across distinct anomaly types and assessing the model’s ability to correctly identify the specific types of anomalies. As previous few-shot anomaly detection methods can not handle the multi-category scenarios, we only compare the baseline model CLIP~\cite{radford2021learningclip} and the vision-language model tailored for medical image MedCLIP~\cite{wang2022medclip}.  In the following, we present the performance of our experiments in the two settings.
\subsubsection{Multi-label Prediction}
For multi-label prediction, we utilize two evaluation metrics to quantify the performance, namely Hamming score and subset accuracy. We provide the details of the two evaluation metrics in the appendix.

\begin{table}[h]
    \centering
    \vspace{1ex}
    \caption{The 1-shot results of the experiments on multi-label prediction. The evaluation metrics are Hamming score and subset accuracy. "SS" is short for "Sign Selection"}\label{tab:multilabel}
    \begin{tabular}{ll|rrrrrr}\toprule
& &\textbf{Clip} &\textbf{MedClip} &\textbf{Ours (no SS)} &\textbf{Ours (full model)} & \\\hline
\multirow{2}{*}{slice 0}&Hamming(\%)$\uparrow$ &80.2 &77.1&85.8 & \textbf{87.2}\\
&Subset acc.(\%)$\uparrow$ &0.4 &18.5 &34.6 &\textbf{60.8}\\\hline
\multirow{2}{*}{slice 5} &Hamming(\%)$\uparrow$ &\textbf{77.6} &63.5& 72.7&76.5   \\
&Subset acc.(\%) &0 &0&\textbf{29.0} &27.3  \\\hline
\multirow{2}{*}{slice10} &Hamming(\%)$\uparrow$ &78.3 &73.2 &73.8 &\textbf{79.2} & \\
&Subset acc.(\%)$\uparrow$ &0 &1.9 &19.8 &\textbf{21.7} & \\
\bottomrule
\end{tabular}
\end{table}
Table \ref{tab:multilabel} shows the 1-shot results with the two evaluation metrics. As shown in the table, our full model demonstrates superior performance compared to the other methods. Also, the comparison between the methods with and without sign selection illustrates the effectiveness of our inference strategy. From the table, we can tell that the pretrained model with vanilla training process can solve the multi-label prediction task for the medical anomaly detection. The comparison between our method without sign selection and vanilla CLIP illustrates the effectiveness of our training process. In addition, the comparison of the performance with and without sign selection also demonstrates the validity of our inference stage.  
\subsubsection{Category-wise AUROC}
\begin{wrapfigure}{r}{0.45\linewidth}
\vskip -0.35in
\begin{center}
\centerline{\includegraphics[width=1.0\linewidth]{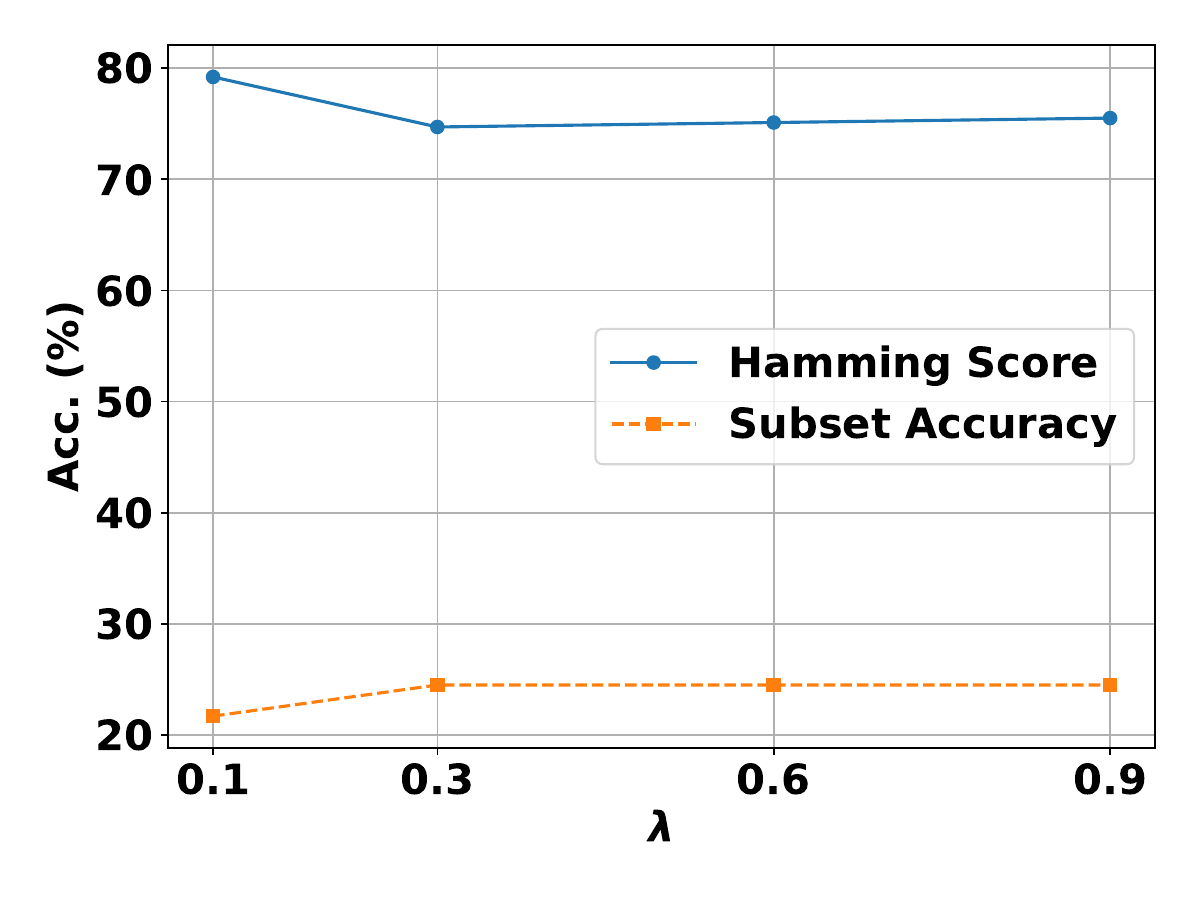}}
\caption{The ablation study on $\lambda$. We conduct the experiments on the multi-label prediction task with two metrics, namely Hamming score and Subset accuracy.}
\label{fig:lamda}
\end{center}
\vskip -0.35in
\end{wrapfigure}
To evaluate the ability to recognize the specific anomaly type, we introduce the category-wise AUROC metric. This metric indicates the performance of the model in each category. Since the label \textit{ Small vessel chronic white matter ischemic change} can not be achieved in slices 5 and 10, we only evaluate 5 categories within these two slices. \par
As Table \ref{tab:categoryauroc} shows, our methods significantly improve the average performance for every slice. Sign selection may not be able to improve category-wise performance. We assume that the reason is that the outlier prompt may fit some images in the test samples. Even if during the training stage, the few-shot samples may underfit these outlier prompts, the prior information may still fit them well to the corresponding anomalies. Thus, the sign selection may not perform well for category-wise settings. The performance on \textit{Enlarged ventricles} can illustrate this issue clearly. 

\begin{table}[h]\centering

\caption{The 1-shot results of the experiments on category-wise AUROC. The reported results are AUROC score (\%).  We also report average (Avg.) results for each slice. "Small vessel ischemic change" corresponds to the label " Small vessel chronic white matter ischemic change" in the FastMRI+ dataset. "SS" and "Avg." are short for "Sign Selection" and "average" respectively.}\label{tab:categoryauroc}
\vspace{1ex}
\resizebox{0.99\textwidth}{!}{
\begin{tabular}{ll|rrrrrr}\toprule
 &&\textbf{CLIP} &
 \textbf{MedCLIP} &\textbf{Ours(no SS)} &\textbf{Ours(full model)} \\\hline
\multirow{7}{*}{slice 0}&Craniotomy &42.7 &50.0 &68.2 &\textbf{70.9}  \\
&Posttreatment change &\textbf{73.5} &51.1 &67.3 &71.7  \\
&Nonspecific lesion &56.8 &44.3 &\textbf{65.1} &56.7  \\
&Dural thickening &44.6& 48.5 &\textbf{58.9} &57.5  \\
&Enlarged ventricles &65.3 &68.7 &62.5 &\textbf{71.9}  \\
&Small vessel ischemic change &62.1& 39.4 &\textbf{81.7} &80.3 & \\\cline{2-7}
&Avg. &57.5 &50.3 &67.3&\textbf{68.2} \\\hline
\multirow{7}{*}{slice 5}&Craniotomy &\textbf{67.2} &46.9 &55.0 &55.4  \\
&Posttreatment change &59.4 &51.7 &\textbf{63.9} &62.4 \\
&Nonspecific lesion &47.9 &44.2 &\textbf{64.9} &\textbf{64.9}  \\
&Dural thickening &51.1& 63.9 &\textbf{64.5} &57.4  \\
&Enlarged ventricles &76.1 &51.3 &\textbf66.6 &\textbf{79.3}  \\\cline{2-7}
&Avg. &60.3 &51.6 &63.0&\textbf{63.9} \\\hline
\multirow{7}{*}{slice 10}&Craniotomy &48.3 &43.0 &51.6 &\textbf{62.4}  \\
&Posttreatment change &\textbf{61.1} &58.2 &40.6 &46.6  \\
&Nonspecific lesion &37.5 &45.6 &\textbf{71.3} &58.0  \\
&Dural thickening &57.7& 48.8 &\textbf{72.2} &69.9  \\
&Enlarged ventricles &98.1 &40.1 &\textbf{100} &70.5  \\
\cline{2-7}
&Avg. &60.5 &47.1 &\textbf{67.1}&61.5 \\
\bottomrule
\end{tabular}}
\end{table}
\subsection{Visualization of Image-Text Similarity}
\begin{figure}
    \centering    
    \begin{subfigure}[t]{0.32\linewidth}
\includegraphics[width=\linewidth]{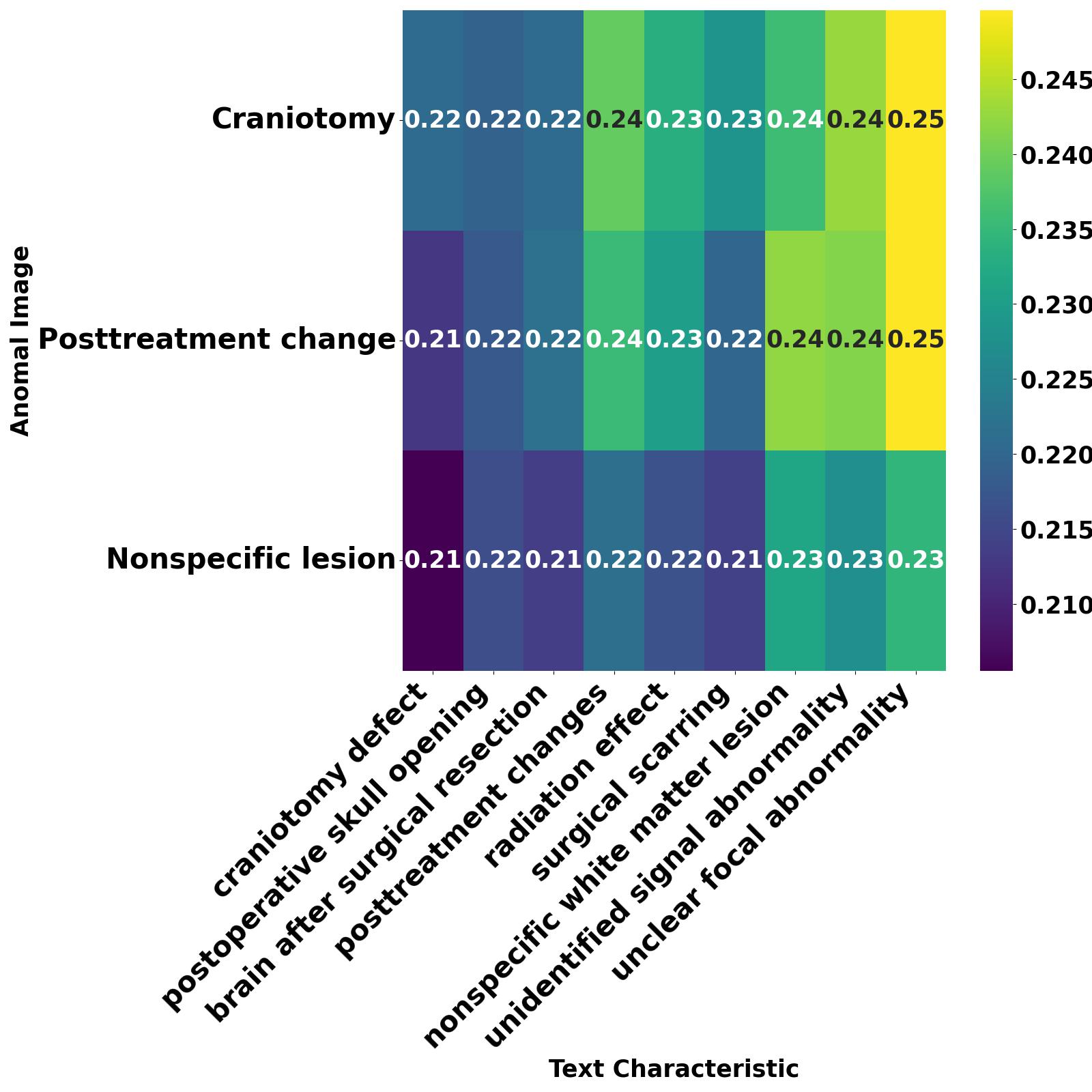}
\subcaption{Heatmap on vanilla CLIP}
    \end{subfigure}
    \begin{subfigure}[t]{0.32\linewidth}
        \includegraphics[width=\linewidth]{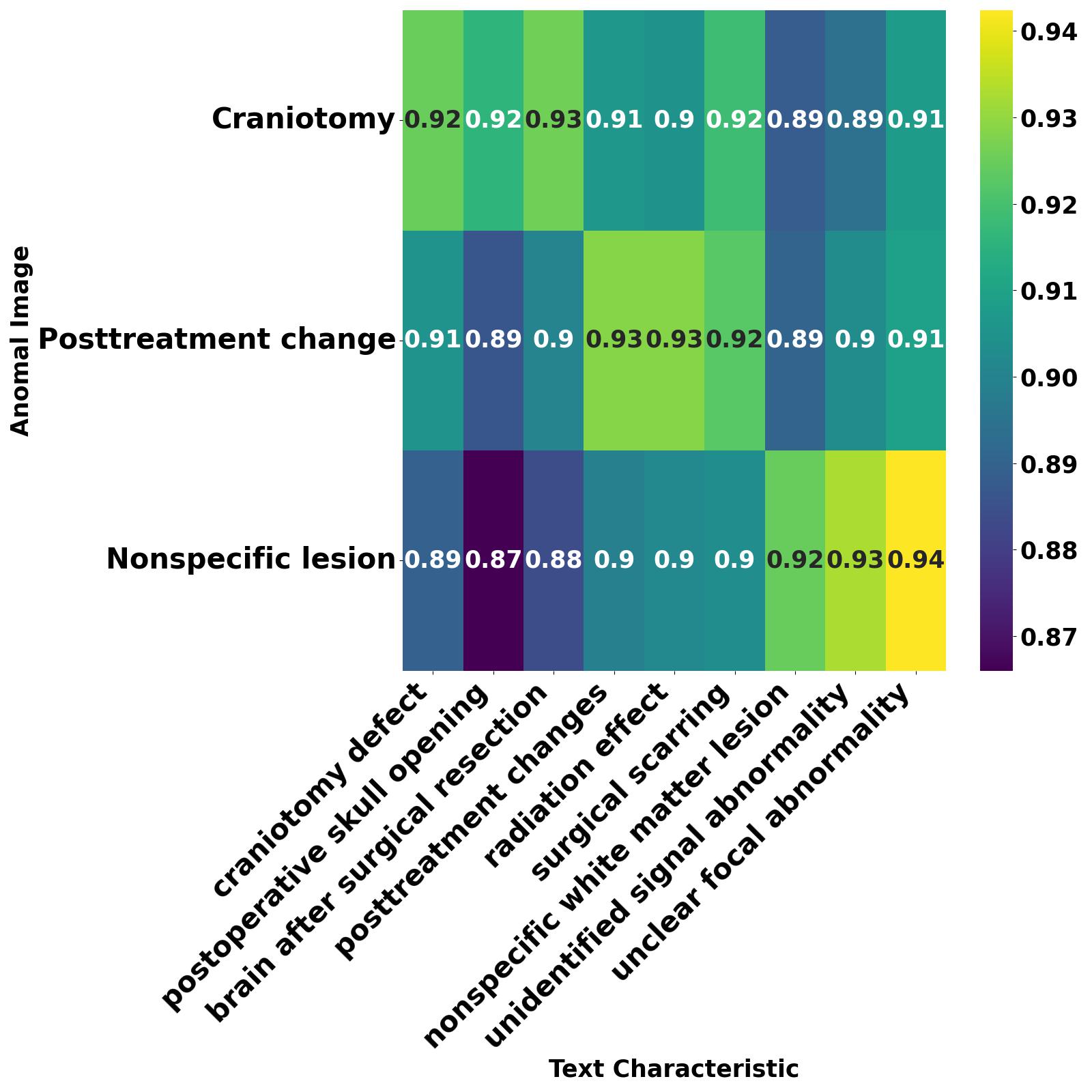}
        
\subcaption{Heatmap on CLIP with our training method}
\label{fig:heatmapb}
    \end{subfigure}
    \raisebox{1cm}{
    \begin{subfigure}[t]{0.32\linewidth}
    \includegraphics[width=\linewidth]{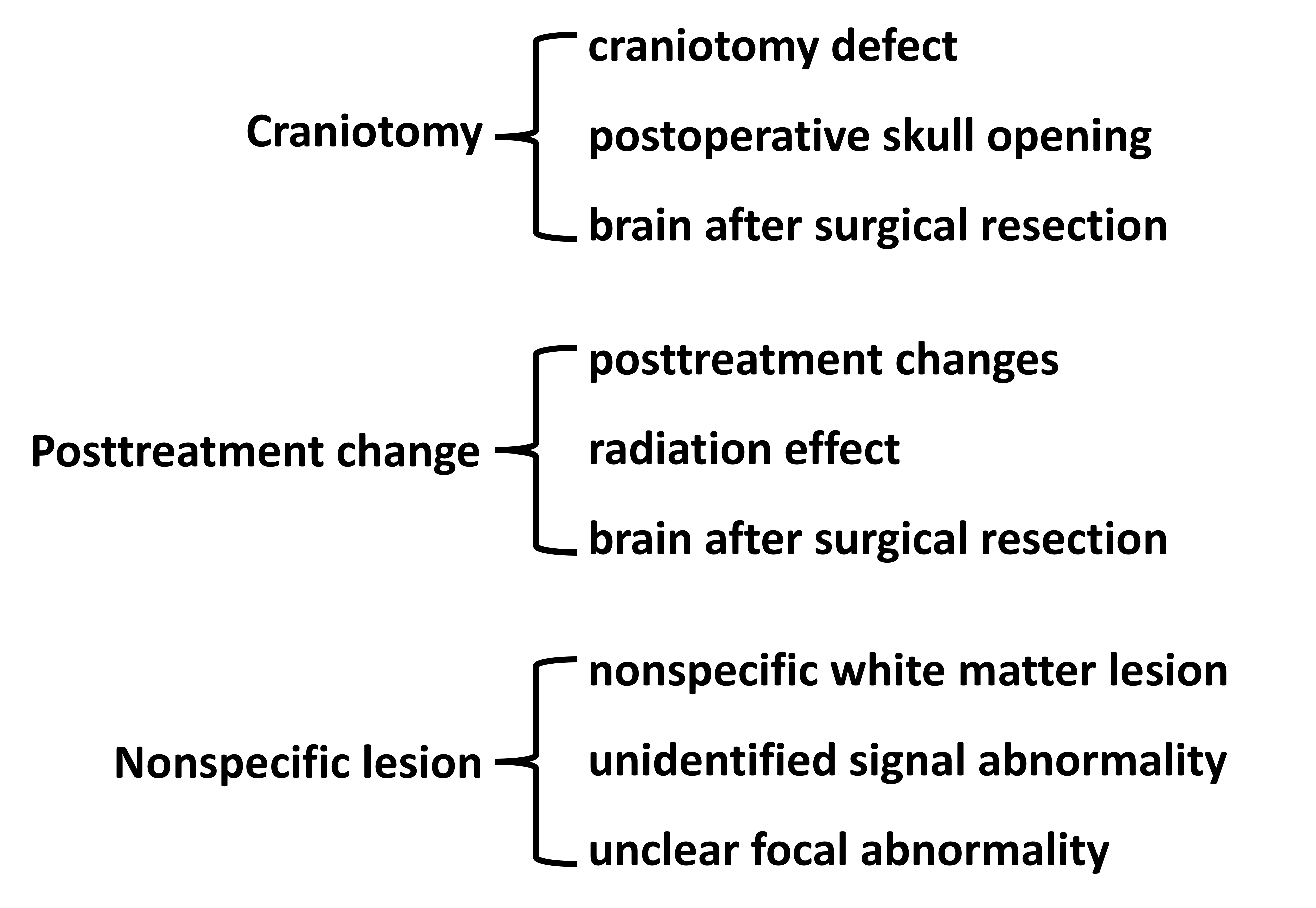}\vspace{1cm}
    \subcaption{Correspondence between prompts and anomaly categories}
    \end{subfigure}}
    \caption{Visualization of image-text similarity heatmaps. (a) visualizes the heatmap on vanilla CLIP, (b) visualizes the heatmap on our trained model. The correspondence between prompts and anomaly categories is provided in (c).}
    \label{fig:heatmap}
\end{figure}

To evaluate the alignment of our method, we visualize the alignment in Figure \ref{fig:heatmap}. As the figure shows, our training method promotes the alignment between abnormal images and corresponding prompts. However, some prompts may exhibit overconfidence in a false category. For instance, in Figure \ref{fig:heatmapb}, the characteristic \textit{surgical scaring} has equally high similarities between both \textit{Craniotomy} and \textit{Posttreatment change} images. This phenomenon may result in uncertain samples for the prediction, which leads us to propose the sign selection method.
\subsection{Ablation study on $\lambda$}
To evaluate the effect of the hyperparameter in the Shift Adapter, we conducted ablation studies for $\lambda$ on the multi-label prediction with the 5th slice. Figure \ref{fig:lamda} shows the experimental results. The figure does not exhibit significant impacts on the performance when $\lambda$ changes, which shows the robustness of the learnable adapter. In addition, we find that there is a trade-off between Hamming score and Subset accuracy when $\lambda$ increases. We assume that it is because the increase of $\lambda$ may cause slight overfitting to the few-shot samples. Therefore, the model may produce fewer predictions in the presence of intra-class variation within the same anomaly type. While this may lead to a reduction in Hamming score, it could potentially enhance the overall prediction accuracy.

\section{Conclusion and Limitation}
In this paper, we introduce a novel setting for medical anomaly detection, termed multi-anomaly detection. Unlike previous settings that typically assume a single anomaly per image, multi-anomaly detection is designed to address scenarios where multiple anomalies co-exist within the same clinical image. Building on this new task, we propose a method based on a vision-language model (VLM) for both inter- and intra-anomaly 
alignment. Specifically, we propose an inter-anomaly loss to amplify the inter-anomaly discrepancy and update the CLIP model with trainable Shift Adapters. In addition, we design a sign selection method to mitigate the intra-anomaly uncertainty at the 
inference stage. To thoroughly evaluate the performance of our method in the task of multi-anomaly detection, besides the general setting in anomaly detection, we propose two more evaluation protocols, namely multi-label prediction and category-wise AUROC. The extensive experiments illustrate the effectiveness of our method. \par
\textbf{Limitation} Even if our proposed method can effectively address the multi-anomaly detection task, there are still limitations, which mainly rely on the correspondence between the prompt and the anomaly categories. Some prompts may correspond to more than one anomaly type, which may result in false predictions if we ignore this nature. Addressing this ambiguity in prompt-anomaly correspondence will be the focus of our future work.

\bibliographystyle{abbrv}
\bibliography{sd_mad}

\clearpage

\end{document}